\documentclass{article} 
\usepackage{iclr2022_conference,times}

\usepackage{amsmath,amsfonts,bm}

\def\eqref#1{equation~\ref{#1}}

\def\1{\bm{1}}

\DeclareMathAlphabet{\mathsfit}{\encodingdefault}{\sfdefault}{m}{sl}
\SetMathAlphabet{\mathsfit}{bold}{\encodingdefault}{\sfdefault}{bx}{n}

\usepackage{hyperref}
\usepackage{url}
\usepackage{enumitem}
\usepackage{makecell}
\usepackage{diagbox}
\usepackage{subfigure}
\usepackage{amsmath, bm}
\usepackage{dsfont}
\usepackage{amssymb}
\usepackage{multirow}
\usepackage{varwidth}
\usepackage{pifont}
\usepackage{makecell}
\usepackage{wrapfig}
\usepackage{multicol}
\usepackage{graphicx} 
\usepackage{comment}
\usepackage{color}
\usepackage{xcolor}
\usepackage{colortbl,booktabs}

\usepackage{algorithmic}
\usepackage[ruled, linesnumbered]{algorithm2e}

\definecolor{Tianlong_color}{rgb}{0.858, 0.188, 0.478}

\title{Spending Your Winning Lottery Better After Drawing It}

\author{%
 Ajay Kumar Jaiswal \textsuperscript{1}, Haoyu Ma\textsuperscript{2}, Tianlong Chen\textsuperscript{1}, Ying Ding\textsuperscript{1}, \textbf{Zhangyang Wang\textsuperscript{1} } \\
  {
  \textsuperscript{1}University of Texas at Austin, 
  \textsuperscript{2}University of California, Irvine 
  } \\
  \small{\texttt{\{ajayjaiswal, tianlong.chen, atlaswang\}@utexas.edu, }} \\
  \small{\texttt{haoyum3@uci.edu, ying.ding@ischool.utexas.edu} } \\

}

\iclrfinalcopy 
\begin{document}

\maketitle

\begin{abstract}

Lottery Ticket Hypothesis (LTH) \citep{frankle2018the} suggests that a dense neural network contains a sparse sub-network that can match the performance of the original dense network when trained in isolation from scratch. Most works retrain the sparse sub-network with the same training protocols as its dense network, such as initialization, architecture blocks, and training recipes. However, till now it is unclear that whether these training protocols are optimal for sparse networks. 

In this paper, we demonstrate that it is unnecessary for spare retraining to strictly inherit those properties from the dense network. Instead, by plugging in purposeful ``tweaks" of the sparse subnetwork architecture or its training recipe, its retraining can be significantly improved than the default, especially at high sparsity levels. Combining all our proposed ``tweaks" can yield the new state-of-the-art performance of LTH, and these modifications can be easily adapted to other sparse training algorithms in general. Specifically, we have achieved a significant and consistent performance gain of $1.05\% - 4.93\%$ for ResNet18 on CIFAR-100 over vanilla-LTH. Moreover, our methods are shown to generalize across datasets (CIFAR10, CIFAR100, TinyImageNet) and architectures (Vgg16, ResNet-18/ResNet-34, MobileNet). All codes will be publicly available.

\end{abstract}

\section{Introduction}


Deep neural networks (NN) have achieved significant progress in many tasks such as classification, detection, and segmentation. However, most existing models are computationally extensive and overparameterized, thus it is difficult to deploy these models in real-world devices. 
To address this issue, many efforts have been devoted to compressing the heavy model into a lightweight counterpart. Among them, network pruning \citep{lecun1990optimal, han2015deep, han2015learning, li2016pruning, liu2018rethinking}, which identifies sparse sub-networks from the dense networks by removing unnecessary weights, stands as one of the most effective methods. Previous methods \citep{han2015learning} usually prune the dense network after the standard training process to obtain the sparse sub-networks. The performance of the pruned network, however, decreases heavily as parts of the weights are removed, and retraining is thus required to recover the original performance \citep{han2015learning}. 

The recent Lottery Ticket Hypothesis (LTH) \citep{frankle2018the} represents a major paradigm shift from conventional after-training pruning. LTH suggests that a dense NN contains sparse subnetworks, named ``winning tickets”, which can match the performance of the original dense NN when trained in isolation from scratch. Such a winning ticket can be ``\textbf{drawn}" by finding the sparse weight mask, from dense training accompanied with iterative magnitude pruning (IMP). The found sparse mask is then applied to the original dense NN, and the masked sparse subnetwork is subsequently re-trained from scratch. Using a similar metaphor, We call the sparse re-training step as ``\textbf{spending}" the lottery, after it is drawn. 

In most (if not all) LTH literature \citep{frankle2018the, frankle2019stabilizing, Renda2020Comparing}, the re-training step takes care of the masked subnetwork, which is re-trained with the same initialization (or rewinding) and same training recipe as its dense network. In plain language, ``\textit{You spend the same lottery ticket in the same way you draw it}". Recent evidence seems to support this convention by attributing LTH's success in recovering the original pruned solution \cite{evci2020gradient}. 
However, till now it is still unclear that whether the architecture blocks, initialization regimes, or training recipes are necessarily optimal for the sparse network. Our question of curiosity is hence: ``\textit{Can you spend the same lottery ticket in a different yet better way than how you draw it}"?




Contrary to the common beliefs, this paper demonstrates that it is unnecessary for sparse network retraining (``spending the lottery") to stick to the protocol of dense network training or sparse mask finding (``drawing the lottery"). Instead, having sparse re-training purposely misaligned in some way from dense training can make the found sparse subnetwork work even better. Specifically, we investigate two possible aspects of modified sparse re-training:
\begin{itemize}
    \item \textit{Architecture tweaking:} after the sparse subnetwork is found, we modify the network architecture by: (a) injecting more residual skip-connections that are non-existent in dense networks; and (b) changing the ReLU neurons into smoother activation functions such as Swish \citep{ramachandran2017searching} and Mish \citep{misra2019mish}. 
    \item \textit{Training recipe tweaking:} when training the (found or modified) sparse subnetwork architecture, we modify the training approach by: (c) changing the ``lottery ticket initialization" by  learned layer-wise scaling; and (d) replacing the one-hot labels with either naive or knowledge-distilled soft labels. 
\end{itemize}
Each idea could be viewed as certain type of \textit{learned smoothening} (we will explain later), and is plug-and-play in the sparse re-training stage of any LTH algorithm. Those techniques can be applied either alone or altogether, and can significantly boost the sparse re-training performance in large models and at high sparsities. Note that all above ``tweaks" only affect the sparse re-training stage (we never alter the found sparse mask), but not the dense training/masking finding stage. In fact, our experiments will show that they boost sparse re-training much more than dense counterparts.

Our contributions can be summarized as:
\begin{itemize}
    \item In contrast to the common wisdom that LTH sparse re-training needs to inherit (masked) network architecture, initialization, and training protocol from dense training, we for the first time demonstrate that purposely re-tweaking them will actually improve the sparse re-training step. That urges our rethinking of the LTH's true value.

    
    \item We investigate two groups of techniques to tweak the sparse subnetwork architecture and training recipe respectively. For the former, we inject new skip connections and replace new activation neurons. For the latter, we re-scale the initialization and soften the labels. Each of the techniques improves sparse re-training (much more than they can help dense counterparts), and altogether they lead to further boosts.
    
    
    \item Our extensive experimental results demonstrate that by plugging these techniques in LTH sparse retraining, we can significantly improve the performance of ``winning tickets" at high sparsity levels and large models, setting the state-of-the-art performance bar of LTH. Furthermore, we show that they can benefit other sparse training algorithms in general, and provide visualizations to analyze their successes. 
\end{itemize}

\section{Background Work}
\vspace{-0.5em}
\subsection{The Lottery Ticket Hypothesis}
\vspace{-0.5em}
The LTH \citep{frankle2018the} implies that initialization is the key for sparse network retraining. Beyond image classification \citep{liu2018rethinking,savarese2020winning,wang2020picking,You2020Drawing,ma2021good,chen2020lottery2}, LTH has also been widely explored in numerous contexts, such as natural language processing \citep{gale2019state,yu2019playing,prasanna2020bert,chen2020lottery1,chen2020earlybert}, reinforcement learning \citep{yu2019playing}, self-supervised learning \citep{chen2020lottery2}, lifelong learning \citep{chen2021long}, generative adversarial networks \citep{chen2021gans,kalibhat2020winning,chen2021ultra}, and graph neural networks \citep{chen2021unified}. 

However, when retraining the sparse network, these works still strictly follow the same training recipe from dense networks. The most recent work \citep{tessera2021keep} reveals that focusing on initialization appears insufficient. Optimization strategies, regularization, and architecture choices also play significant roles in sparse network training. 
However, \citep{tessera2021keep} only compares sparse networks to their equivalent capacity dense networks, and most of their experiments are conducted on multi-layer perceptron (MLP). Thus, it is unclear whether their conclusion can generalize to the ticket finding from dense networks.  

\vspace{-0.5em}
\subsection{Smoothness}
\vspace{-0.5em}
Introducing smoothness into NNs, including on the weights, logits, or training trajectory, is a common techniques to improve the generalization and optimization  \citep{jean1994weight}.  
For labels, smoothness is usually introduced by replacing the hard target with soft labels \citep{szegedy2016rethinking} or soft logits \citep{hinton2015distilling}. This uncertainty of labels helps to alleviate the overconfidence \citep{hein2019relu} and improves the generalization.  
Smoothness can also implemented by replacing the activation functions \citep{misra2019mish, ramachandran2017searching}, adding skip-connections in NNs \citep{He2016DeepRL}, or averaging along the trajectory of gradient descent \citep{izmailov2018averaging}. These methods contribute to more stable gradient flows \citep{tessera2021keep} and smoother loss landscapes, but most of them have not been considered nor validated on sparse NNs. 
%


\begin{figure*}
\includegraphics[width=14cm]{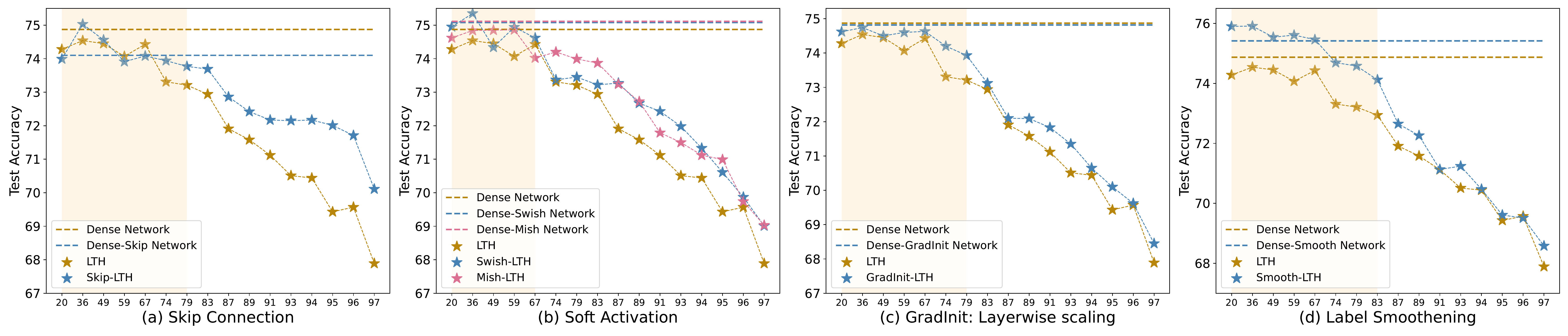}
\caption{\small{Results of testing accuracy for ResNet-18 on CIFAR100 for our architecture tweaking and training recipe tweaking techniques. X-axis denotes the sparsity level of the network. Dash straight lines denote the accuracy of dense network trained with and without proposed tweaking techniques. Performance gain by our techniques is more profound for sparse NN in comparison to dense NN. Shaded part of the graph illustrates the "Winning Tickets" identified by LTH with tweaking techniques. All experiments are conducted using \textit{Iterative Magnitude Pruning} (IMP).}}
\label{fig:methods}
\vspace{-1em}
\end{figure*}

\begin{wrapfigure}{r}{6cm}
\vspace{-1.2cm}
\includegraphics[width=6cm]{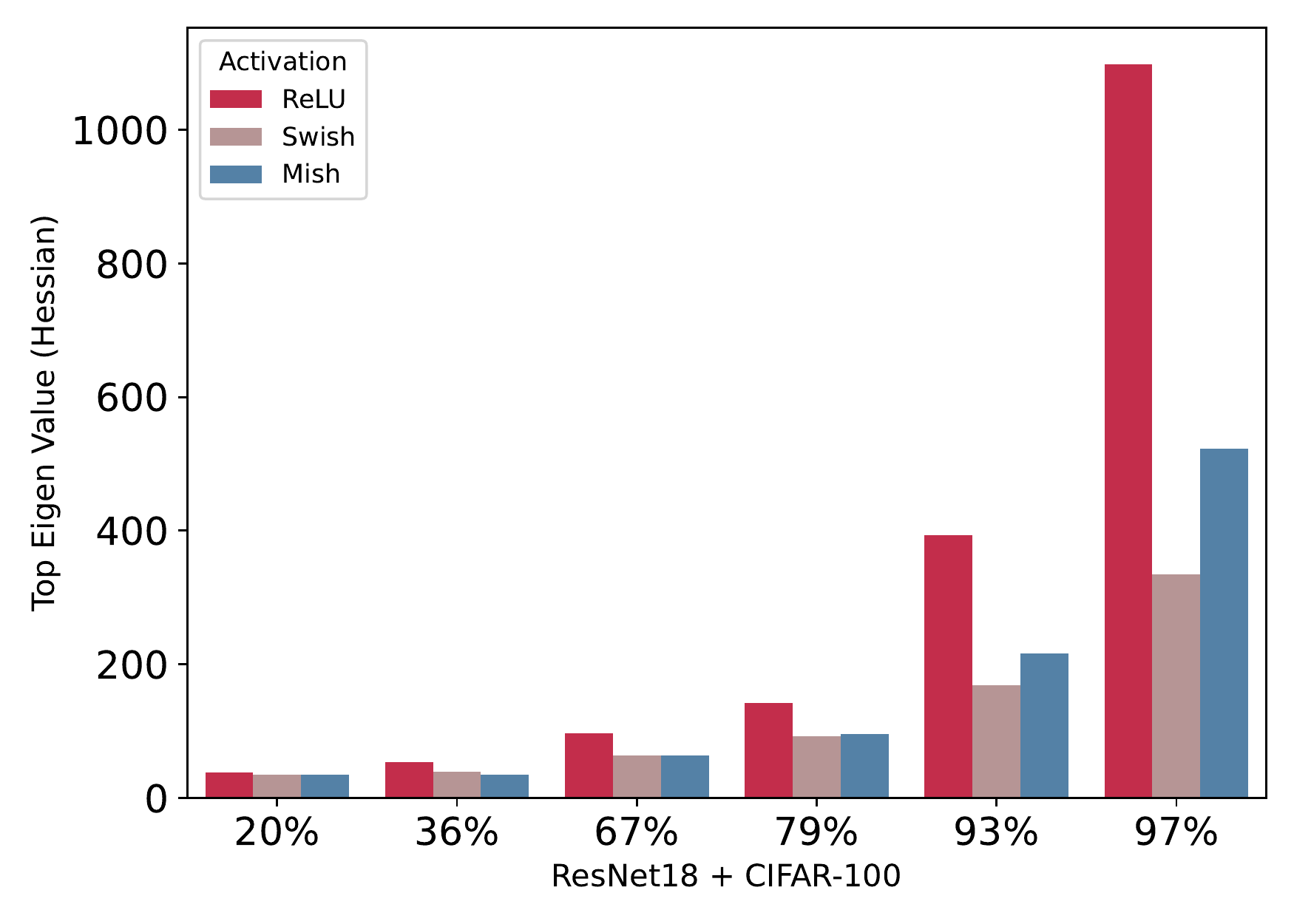}
\vspace{-0.8cm}
\caption{Comparison of the top eigenvalue of Hessian (mean across 10 random batches) of sparse NN trained with different activation functions. Comparatively smaller eigenvalues of soft-activation (Swish/Mish) based sparse NN wrt. ReLU based networks indicate the presence of high smoothness. 
}
\vspace{-1cm}
\label{fig:soft_hessain}
\end{wrapfigure} 

\section{Methodology}

The LTH \citep{frankle2018the} suggests that the ``winning tickets" can be found via the following three steps: (1) training a dense network from scratch; (2) pruning unnecessary weights to form the mask of a sparse sub-network; and (3) re-training the sub-network from the same initialization used in the dense model. For the third step, the retraining of sparse network usually inherits all properties, such as architecture blocks and training recipes, from dense networks. However, those are not necessarily optimal for training sparse networks. In view of that, we investigate two aspects of ``tweaks" dedicated to the sparse re-training step: model architecture, and training recipe.

\subsection{Model Architecture tweaking}

\paragraph{Replacing Smoother Activations} 

Most deep NNs apply the Rectified Linear Units (ReLU) \citep{nair2010rectified} as the activation function. However, the gradient of ReLU changes suddenly around zero, and this non-smooth nature of ReLU is an obstacle to sparse retraining because it leads to high activation sparsity into the subnetwork, likely blocking the gradient flow. To mitigate this issue and encourage healthier gradient flow, we replace the ReLU to Swish \citep{ramachandran2017searching} and Mish \citep{misra2019mish}. Different from ReLU, Swish and Mish are both smooth non-monotonic activation functions. The non-monotonic property allows for the gradient of small negative inputs, which leads to a more stable gradient flow \citep{tessera2021keep}. Meanwhile, the loss landscape of Swish and Mish are proved to have smoother transition \citep{misra2019mish}, which makes the loss function easier to optimize and hence makes the sparse network generalize better.

\paragraph{Injecting New Skip Connections}
High sparsity networks easily suffer from the layer-collapse \citep{tanaka2020pruning}, i.e., the premature pruning of an entire layer. This could make the sparse network untrainable, as the gradient cannot be backpropagated through that layer. 
The skip connection ( or named "residual-addition" ) \citep{He2016DeepRL} was initially proposed to avoid gradient vanishing problem, and enables the training of a very deep model. It is later proven that skip connections can smooth the loss surfaces \citep{li2017visualizing}. That naturally motivates us to consider using more skip connections on the sparse networks to smoothen the landscape and preserve gradients, besides its possible mitigation effect when encountering collapsed layers. 

\begin{figure}
\vspace{-0.7cm}
\centering
\includegraphics[width=12cm]{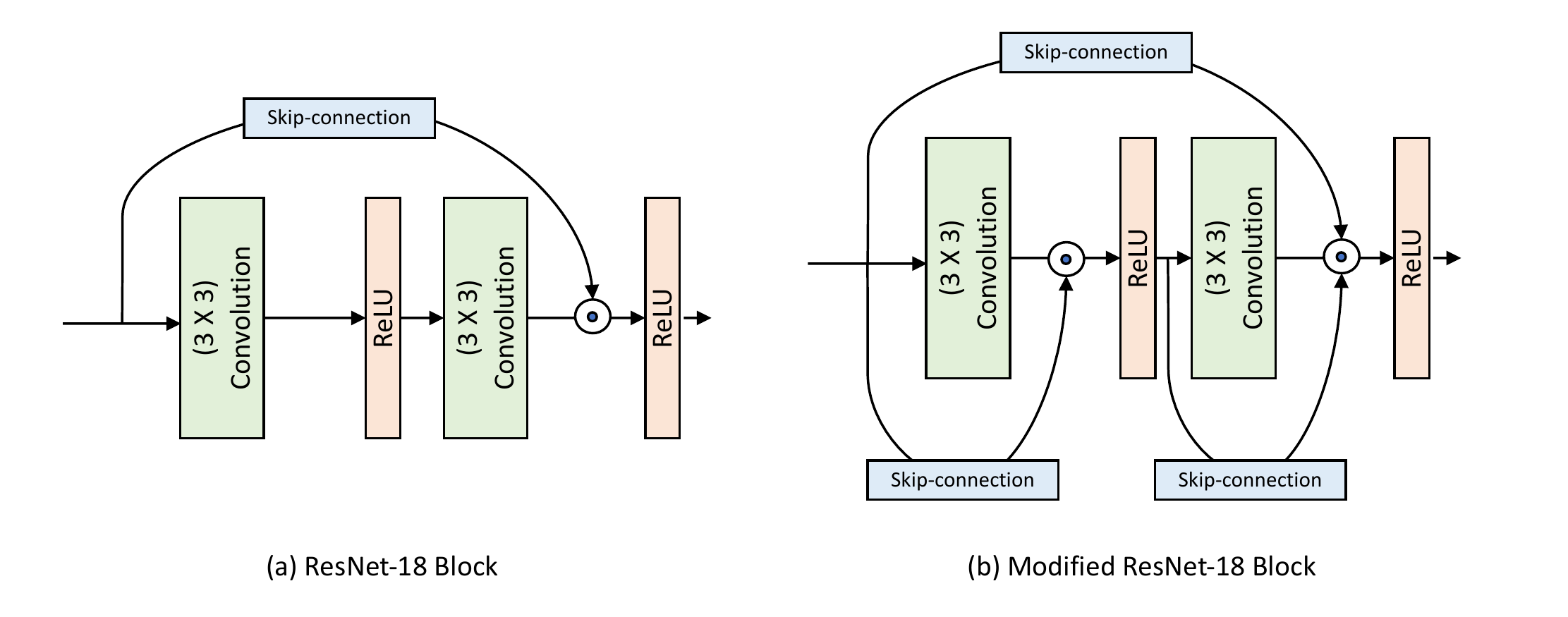}
\vspace{-0.5cm}
\caption{Sketch of our modified ResNet-18 block to introduce additional skip-connections for sparse re-training. }
\vspace{-0.3cm}
\label{fig:skip_model}
\end{figure}

Inspired by \citep{ding2021repvgg}, we propose to ``artificially" add new skip-connections during our sparse re-training. Figure \ref{fig:skip_model} illustrates this architectural modifications to the traditional Resnet-18 block. Similar to existing residual connection in traditional ResNet-18 block, our newly introduced skip-connections add input of each $(3 \times 3)$ convolution block, to their output before the activation. 
The original motivation behind residual connections comes from their ability to allow gradient information to flow to earlier layers of the NN, thereby reducing the
vanishing gradient problem during training \citep{He2016DeepRL}. With high activation sparsity present in sparse subnetworks, additional skip-connections can facilitate healthy gradient flow and improve their trainability. Furthermore, \citep{li2017visualizing} observed that with an increase in depth of networks, neural loss landscape becomes highly chaotic and leads to drop in generalization and trainability. They further observed that skip connections promote flatness and prevent transition to chaotic behaviour. Inspired by them, we added skip-connections to stabilize the initial chaotic training regime of sparse NN at high sparsity level and manage to prevent the transition to a sub-optimal behaviour.

\subsection{Training recipe tweaking}

\paragraph{Smoother Labels and Loss functions}
Label smoothing \citep{szegedy2016rethinking} has been widely used to improve the performance of dense networks trained with hard labels. 
Specifically, given the output probabilities $p_{k}$ from the network and the target $y_{k}$, a network trained with hard labels aims to minimize the cross-entropy loss by $L_{\text{LS}}= - \sum_{k=1}^{K} y_{k} \log \left(p_{k}\right)$, where $y_{k}$ is "1" for the correct class and "0" for others, and $K$ is the number of classes.  
Label smoothing changes the target to a mixture of hard labels with a uniform distribution, and minimizes the cross-entropy between the modified target $_{k}^{L S}$ and output $p_{k}$. The modified  target is defined as $y_{k}^{L S}=y_{k}(1-\alpha)+\alpha / K$, where $\alpha$ is the smooth ratio.  
This uniform distribution introduces smoothness into the training and encourages small logit gaps. Thus, label smoothing results in better model calibration and prevents overconfident predictions \citep{muller2019does}. 

Meanwhile, recent works \citep{yuan2020revisiting, shen2021label} suggest that Knowledge Distillation (KD) \citep{hinton2015distilling} is also one kind of label smoothing. KD aims to make (student) models learn from outputs produced by pretrained teacher models. 
In detail, given $q_k^{\tau}$ and $p_k^{\tau}$, the outputs from both teacher and student after softmax with temperature $\tau$, respectively, the network trained with KD aims to minimize  $L_{\text{KD}} = \mathcal{KL} (q_k^{\tau}, p_k^{\tau})$, where  $\mathcal{KL}$ is the Kullback-Leibler divergence. Compared with uniform distribution, the outputs from teachers assign incorrect classes with different probabilities, which reveals a rich similarity structure over the data \citep{hinton2015distilling}. 

Both loss functions have been proved beneficial in dense network training \citep{yuan2020revisiting}. 
However, till now it is unclear whether they are still helpful in sparse network training. We demonstrate that their smoothness property can bring additional benefits for sparse network training, as they can flatten the loss landscapes and facilitate the training. We also observe that naive label smoothening works sufficiently well compared to KD: please refer to Section 4.6.


\paragraph{Layer-wise Re-scaled Initialization}

Carefully crafted initialization techniques that can prevent gradient explosion/vanishing in backpropagation have been an important part of the early success of feed-forward neural networks \citep{He2016DeepRL, Glorot2010UnderstandingTD}. Even with recent cleverly designed initialization rules, complex models with many layers and branches suffer from instability. For example, the Post-LN Transformer \citep{Vaswani2017AttentionIA} can not converge without learning rate warmup using the default initialization. 

In sparse subnetwork re-training, most of the existing works use common initialization methods \citep{Glorot2010UnderstandingTD, He2016DeepRL} derived from dense NN (with the sparse mask applied). These initialization techniques ensure that the output distribution of every neuron in a layer is of zero-mean and unit variance by sampling a Gaussian distribution with a variance based on the number of incoming/outgoing connections for all the neurons in a dense layer. However, after sparsification, the number of incoming/outgoing connections is not identical for all the neurons in the layer \citep{Evci2020GradientFI} and this raises direct concerns against the blind usage of dense network initialization for sparse subnetworks. Furthermore, \citep{Evci2020GradientFI} showed that completely random re-initialization of sparse subnetworks is also not viable. As the LTH initialization and mask are entangled, fully re-initializing will have us ``lose the drawn lottery", resulting in the retrained sparse subnetwork converging to different (usually much poorer) solutions. 

To balance between these concerns, we keep the original initialization intact for each parameter block and just re-scaled it by a learned scalar coefficient following recently proposed in  \citep{Zhu2021GradInitLT}. Aware of the sensitivity and negative impact of changing initialization identified by \citep{evci2020gradient}, we point that that linear scaling will \textbf{not} hurt the original LTH initialization, thanks to the BatchNorm layer which will effectively absorb any linear scaling of the weights.  More specifically, we optimized a small set of scalar coefficients to make the first update step (e.g., using SGD) as effective as possible at lowering the training loss. After the scalar coefficients are learned, the original LTH initializations of subnetworks are re-scaled and the optimization proceeds as normal.

\begin{figure}
\centering
\includegraphics[width=14cm]{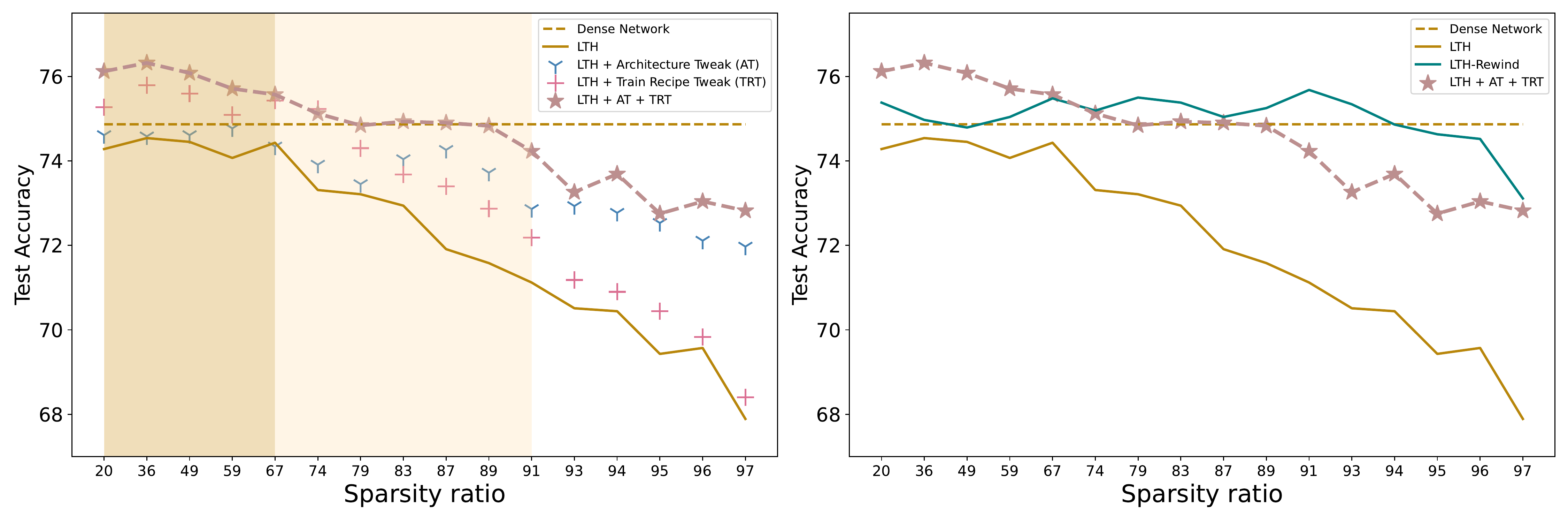}
\vspace{-0.4cm}
\caption{(a) Comparison of our techniques (architecture tweaking and training recipe tweaking) with respect to vanilla LTH using ResNet-18 on CIFAR100. We observe significant improvements in the performance of sparse networks when we combine our techniques during the \textit{spending} process of tickets. "Winning tickets" can be identified for sparsity as high as 91\% for ResNet-18 on CIFAR100. (b) Comparison of test accuracy of LTH trained with our techniques and LTH-Rewind.}
\label{fig:combining}
\end{figure}

\subsection{Analysis: Commodity of our ``tweaks"}

In this section, we try to understand the implications of our techniques during training of sparse subnetworks, through some common lens. We study the loss landscape contours of our proposed techniques during the early training phase and late training phase and compare them with vanilla-LTH. Our methods can be viewed as a form of \textit{learned smoothening} which can be incorporated at an early or late stage of the training. Smoothening tools can be applied on the logits (naive label-smoothening, knowledge distillation), on the weight dynamics (stochastic weight averaging), or on regularizing end solution. Sparse subnetworks trained from scratch suffer from high activation sparsity due to gradient clipping in the negative range of ReLU and it can be mitigated by replacing ReLU with smooth activations (Appendix \ref{tab:activation_sparsity}). Figure \ref{fig:soft_hessain} represents the top eigenvalue of the Hessian of model trained with different activation functions. Top eigenvalue of Hessian can act as a proxy for smoothness \citep{Yao2020PyHessianNN}, and it is evident that ReLU-based sparse networks have comparatively high eigenvalue than smooth activation based subnetworks. 

\begin{figure}
\centering
\includegraphics[width=14cm]{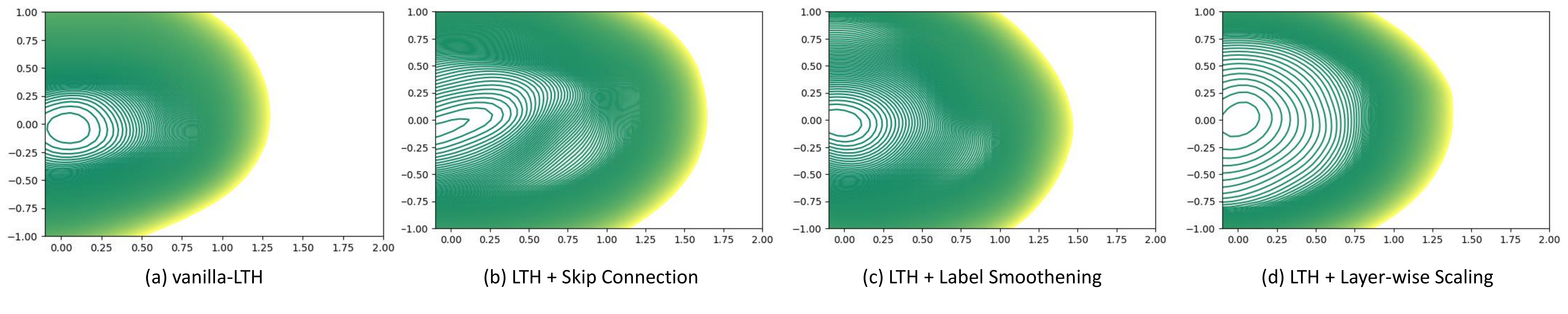}
\vspace{-0.8cm}
\caption{Comparison of loss surface contours of ResNet-18 models with 91\% sparsity in early training stage (epoch 5) using vanilla-LTH and our Smoothness-aware techniques.}
\label{fig:init_analysis}
\vspace{-0.4cm}
\end{figure}

Other three of our proposed techniques smooth different angles of the sparse training. Figure \ref{fig:init_analysis} presents the comparison of the loss contours of ResNet-18 models (91\% sparsity) in their very early training stage (epoch 5) on CIFAR-100. Figure \ref{fig:loss_landscape} presents the loss landscape visualization of the trained models (epoch 180). It can be clearly observed that our architecture changes (skip-connection) can remarkably change the loss landscape (different counter shape, larger landscape area with different basin shape and size in middle) from beginning to end (Figure \ref{fig:init_analysis}(b), \ref{fig:loss_landscape}(c)). As expected, our layer-wise scaling technique also has a very high impact in the early training stage (Figure \ref{fig:init_analysis}(d)). We can observe a high difference in the variance of its contours compare to vanilla-LTH indicating the presence of smoothness. However, label-smoothening tends to impact the later phase of training more (highest difference in loss landscape, Figure \ref{fig:loss_landscape}(d)) compared to the early phase (minimal difference wrt. vanilla-LTH, Figure \ref{fig:init_analysis}(c)).

\section{Experiments and Analysis}
\vspace{-0.5em}
\subsection{Settings}
\vspace{-0.5em}
\paragraph{Datasets and Architecture:} 
Following previous works of LTH \citep{frankle2018the, frankle2019stabilizing}, we  consider four popular  architectures, ResNet-18,  ResNet-34 \citep{He2016DeepRL}, VGG-16 \citep{simonyan2014very} and MobileNet \citep{howard2017mobilenets} to evaluate the effectiveness of our proposed techniques. We considered three datasets in our experiments:  CIFAR-10, CIFAR-100~\citep{krizhevsky2009learning},  and Tiny-ImageNet and reported the performance of our techniques. In all our experiments, we randomly split the original training dataset into one training and one validation sets with a 9:1 ratio. Primary results, ablation, and visualizations are mainly performed on CIFAR-100 using ResNet-18.

\paragraph{Training and Evaluation Details:} For all experiments, we by default use ResNet-18 and CIFAR-100 except except our extensive evaluation across datasets and architectures in Figure \ref{fig:data_model}. For training both the dense and sparse network, we adopt an SGD optimizer with momentum $0.9$ and weight decay $2$e$-4$. The initial learning rate is set to $0.1$, and the networks are trained for 180 epochs with a batch size of 128. The learning rate decays by a factor of $10$ at the $90$th and $135$th epoch during the training. The same training configurations are applied across all the other datasets and models evaluation in Figure \ref{fig:data_model}. For weight rewinding experiments, we rewind the dense network weight to the 33rd epoch to initialize the sparse subnetwork.

\subsection{Architecture Tweaking (AT) in LTH} 
We modify the sparse subnetwork found during the pruning process by introducing additional non-existing residual connections and changing the ReLU neurons with smoother activation functions such as Swish/Mish. Figure \ref{fig:methods}(a), (b) show the performance improvement of our AT techniques in comparison of vanilla-LTH \citep{frankle2018the}. 
Interestingly, the additional skip-connections hurts the performance of the dense network by $-0.77\%$, but it significantly improves the performance of sparse networks, especially at extremely high sparsity levels, such as: $91\% (+1.05\%)$, $95\%(+1.58\%),$ and $97\%(+2.22\%)$. Similarly, the impact of smooth activation functions is marginal for dense networks: swish ($+0.21\%$) and mish ($+0.24\%$), but they also provide significant improvements in sparse network training. Figure \ref{fig:combining}(a) shows the performance when both AT techniques (skip connection + swish activation) are jointly applied in the sparse re-training step of LTH. We can observe that our techniques outperform vanilla-LTH markedly and the performance gap increase with the increase in sparsity level ($+4.08\%$ for sparsity level $97\%$).

\subsection{Training Recipe Tweaking (TRT) in LTH}
Sparse subnetwork training in LTH conventionally derives its training properties from its dense counterpart. We modify the training recipe of sparse subnetworks by changing their initialization \citep{gradinit} and using soft labels instead of one-hot labels. Due to minor performance differences between knowledge-distilled soft labels and naive label smoothening for sparse subnetworks, and the high intertwine of KD-labels with dense training, we choose to conduct our experiments with naive soft labels.   

Figure \ref{fig:methods} (c), (d) show the performance improvement of our training recipe tweaking techniques in comparison of vanilla-LTH \citep{frankle2018the}. We can clearly observe that our techniques benefit sparse subnetworks significantly more than the dense network (dash straight line). Furthermore, they help "winning tickets" identified by LTH to win better, i.e have performance closer to the dense network. For example, winning tickets at 36\% and 59\% sparsity perform $+1.37\%$ and $+1.54\%$ better than vanilla-LTH due to label smoothening. Similarly, winning tickets at 59\% sparsity perform $+0.53\%$ better than vanilla-LTH due to layer-wise scaled initialization. Figure \ref{fig:combining}(a) show the performance when our methods (scaled initialization + soft labels) are jointly applied in the sparse re-training step of LTH. It is evident from the figure that in the range of winning tickets, our methods significantly boost the performance of sparse subnetworks allowing them to perform even better than the dense network. 

\begin{figure}
\centering
\includegraphics[width=14cm]{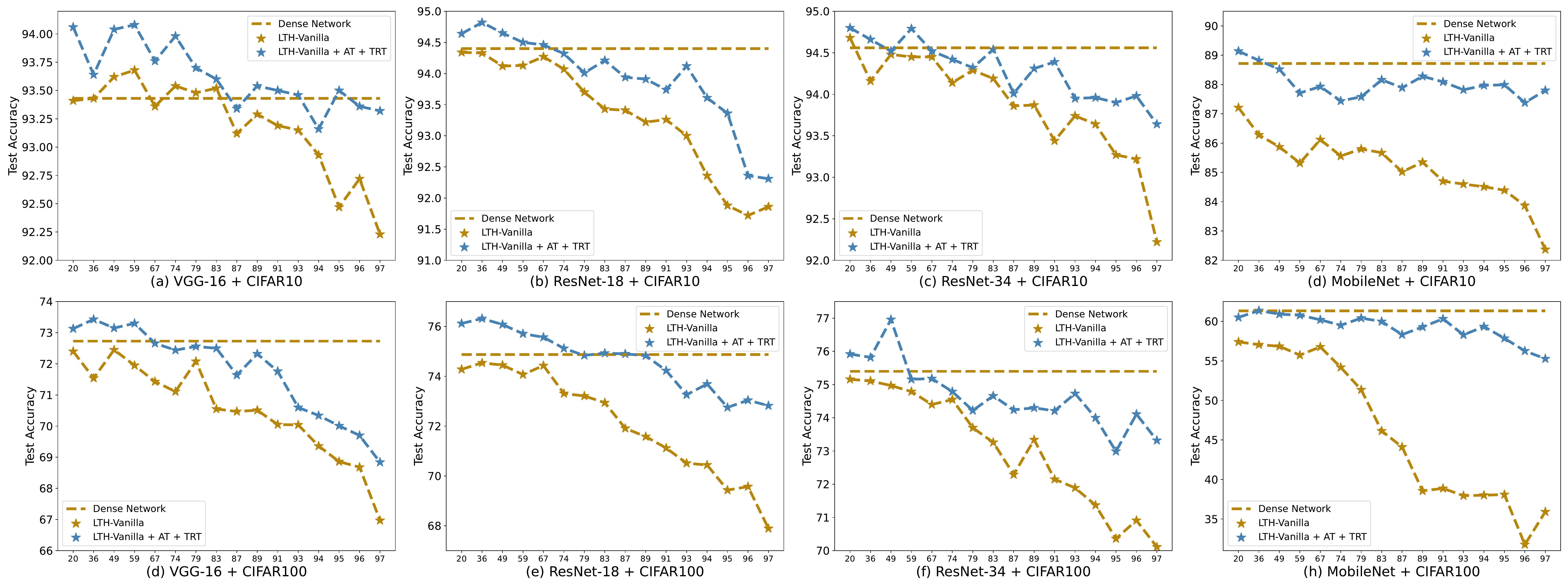}
\caption{Comparison of testing accuracy of proposed techniques (AT + TRT combined) for Vgg16, ResNet-18, ResNet-34, and MobileNet on CIFAR-10, CIFAR-100 datasets wrt. vanilla-LTH. Straight dash line represent the performance of dense network without any tweaking. }
\label{fig:data_model}
\end{figure}

\subsection{Combining AT and TRT}

\begin{wrapfigure}{r}{5.5cm}
\vspace{-1.6cm}
\includegraphics[width=5.5cm]{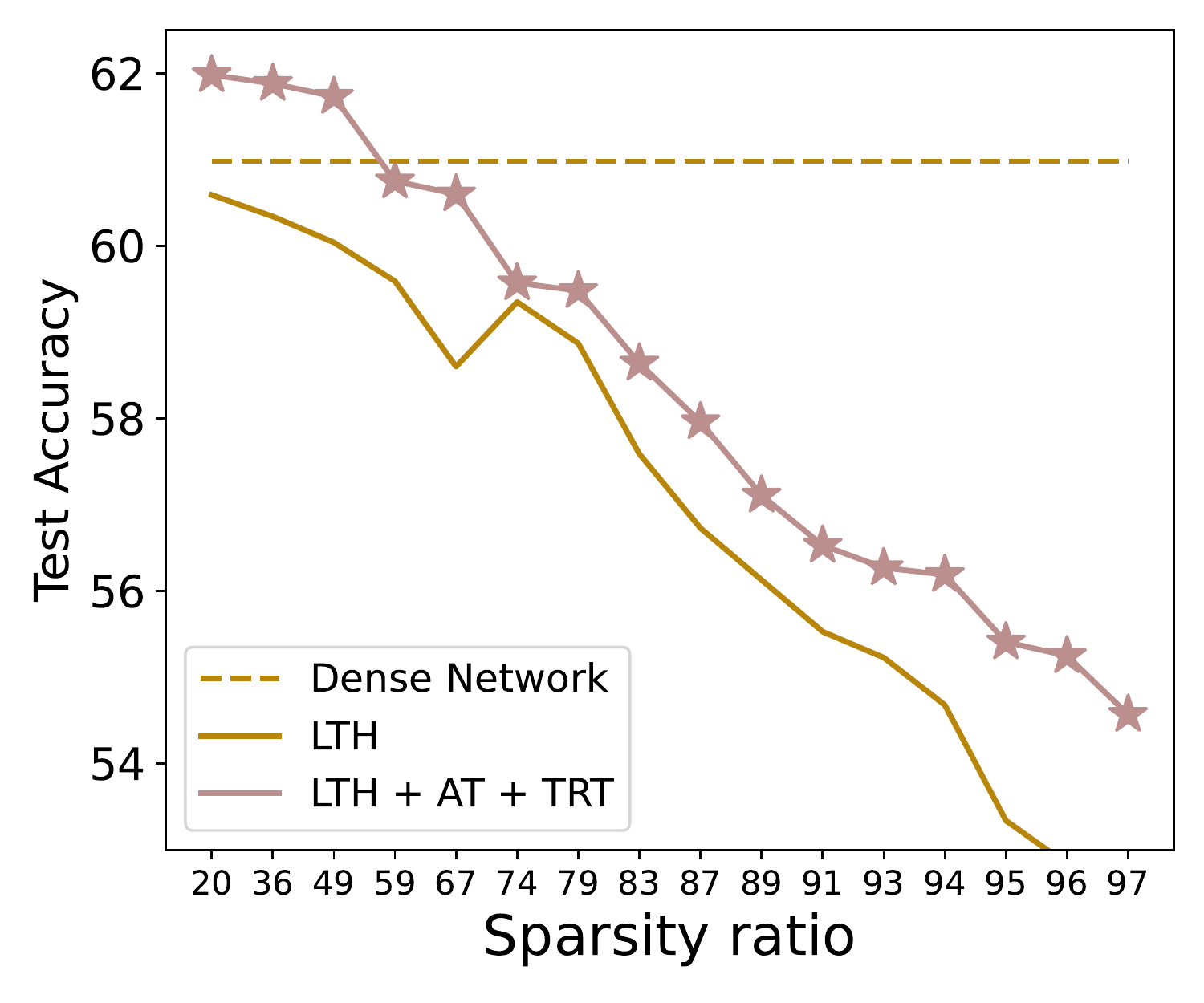}
\vspace{-0.9cm}
\caption{Testing performance of our proposed technique on Tiny-ImageNet.}
\vspace{-0.4cm}
\label{fig:tinyimagenet}
\end{wrapfigure} 

We combine our architecture (AT) and training recipe (TRT) tweaking techniques to train sparse subnetworks. Figure \ref{fig:combining}(a) shows the performance of our combined approach wrt. to individual and vanilla-LTH. Very interestingly, it can be observed that while AT techniques provide significant improvements at high sparsity levels, TRT techniques provide significant improvements at low sparsity level. Due to the orthogonal benefits provided by AT and TRT methods, when they are combined, they patently outperform vanilla-LTH. We observe a huge performance improvement of $+4.93\%$ at sparsity level 97\%. AT + TRT technique allow tickets identified at sparsity as high as 91\% to be "winning tickets", in comparison to 67\% for the vanilla-LTH.

Figure \ref{fig:data_model}, and \ref{fig:tinyimagenet} demonstrate the performance of our proposal (AT + TRT) across ResNet-18,  ResNet-34 \citep{He2016DeepRL}, VGG-16 \citep{simonyan2014very} and MobileNet \citep{howard2017mobilenets} on CIFAR-10, CIFAR-100, and Tiny-ImageNet. We compare our proposal with vanilla-LTH and found that our proposal out-performance prevails across all model architecture and datasets. Due to its simplicity and generalization capability across architectures and datasets, our proposal makes a strong case to be widely accepted for sparse subnetwork training.

\paragraph{Comparison with Weight rewinding:} Weight rewinding \citep{frankle2019stabilizing} proposes to "roll back" the dense model weights to some early training iteration, and use the rewound weight to initialize sparse subnetworks during the re-training step. While this technique has proven to be very helpful for stabilizing sparse re-training and attaining state-of-the-art performance for sparse subnetworks, it has its own drawbacks of the struggle to find rewind point, bookkeeping the weights of dense network during their training, entwine of rewind point to model architecture, etc. Figure \ref{fig:combining}(b) illustrates the performance comparison of AT + TRT (rose brown) with respect to weight rewinding technique (green). In order to compare the performance difference between them, we name the sparsity level where both lines cross each other as \textit{critical sparsity}. At sparsity lower than it, AT + TRT generally outperforms weight rewinding. Furthermore, at some higher sparsity (74\% - 89\%, and 97\%), AT + TRT can still perform comparably with rewinding.

\begin{figure}
\centering
\includegraphics[width=14cm]{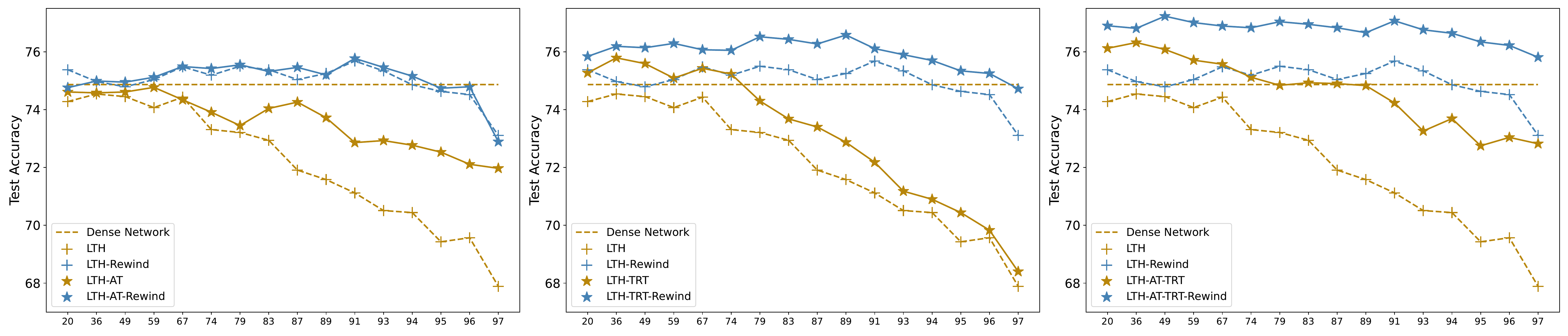}
\vspace{-0.5em}
\caption{Comparison of the testing accuracy of our proposed techniques: \textit{architecture tweaking (AT)} and \textit{training recipe tweaking (TRT)} with respect to \textit{rewinding}. We also show the effect of combining rewinding with AT and TRT. State-of-the-art results are obtained for sparse retraining when our approaches are combined with rewinding.}
\vspace{-1em}
\label{fig:rewind_compare}
\end{figure}

\subsection{Combining AT and TRT with Weight Rewinding: New State-of-the-art}
\begin{wrapfigure}{r}{5.7cm}
\vspace{-0.5cm}
\includegraphics[width=5.7cm]{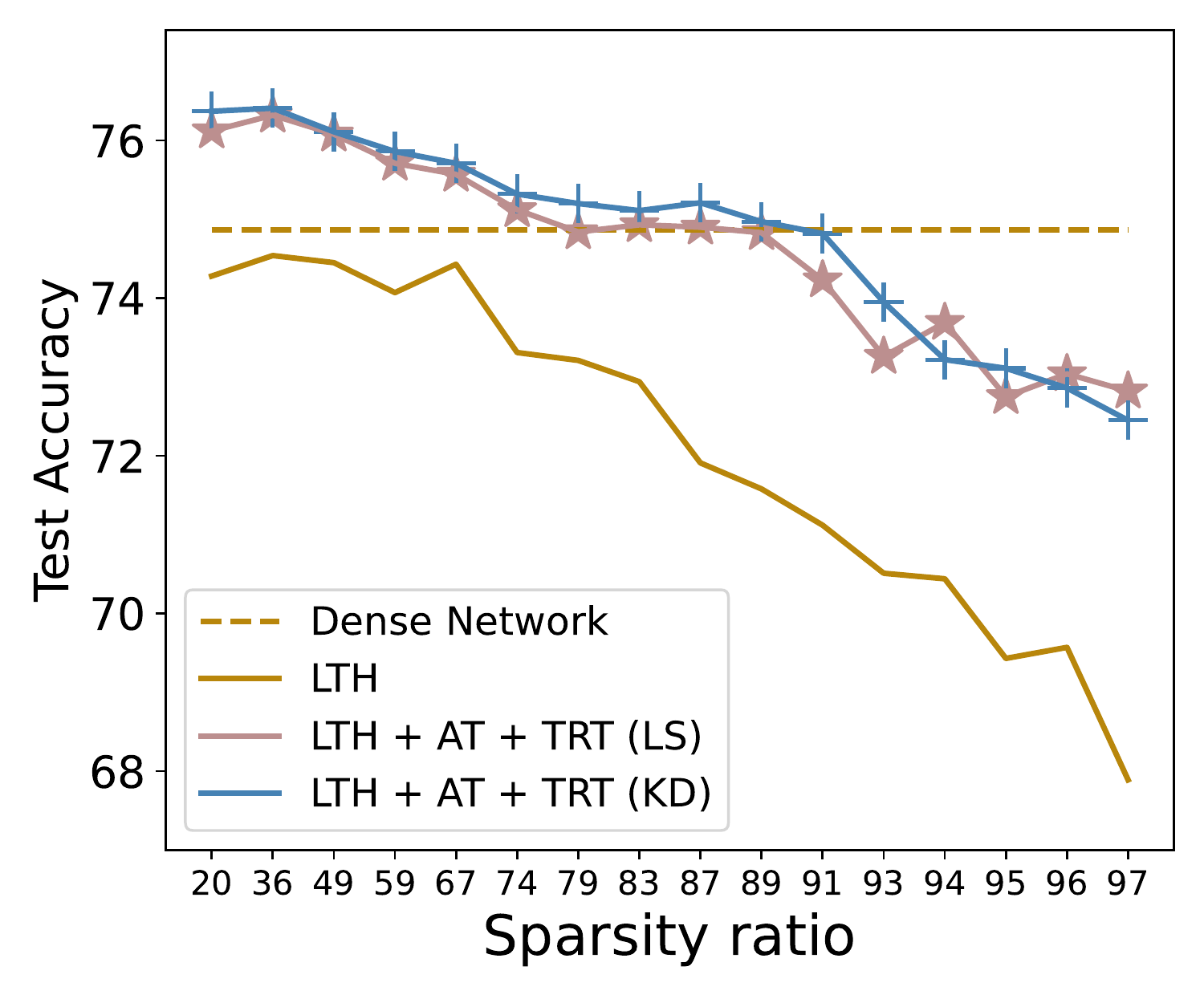}
\vspace{-0.9cm}
\caption{Comparison of our proposed LTH (+AT and +TRT), when naive soft labels (LS) are replaced with knowledge-distilled (KD) soft-labels in TRT.}
\label{fig:LSvsKD}
\end{wrapfigure} 
In this section, we study the combination of our techniques: AT and TRT with weight rewinding. Our AT techniques provide significant improvement at high sparsity, while our TRT techniques help subnetworks with low sparsity. Being orthogonal, when they are combined, their performance is very comparable to weight rewinding. To further investigate how rewinding can improve the performance of our proposal, we perform experiments to combine both our proposal and weight rewinding technique together. In detail, we rewind the remaining weights of our pruned subnetwork to a specific early training point (18\% in our experiments).

Figures \ref{fig:rewind_compare} presents the results of our proposed techniques when they are combined with weight rewinding. We show the performance improvement with rewinding individually for AT and TRT, as well as when they are combined together. We observe that AT + rewinding performs marginally better than rewinding, but TRT + rewinding performs significantly better than rewinding. However, when AT + TRT together is combined with weight rewinding, we achieve state-of-the-art performance. Very interestingly, we observe a very marginal impact of sparsity on the performance. The performance of subnetworks does not vary much with increase in compression level. Overall performance is around $>2.0\%$ than the dense network for almost all sparsity levels. The success of our proposed techniques with minimal changes in the current LTH paradigm, sets the new benchmark for sparse re-training and validates that sparse re-training needs not to intertwine with dense training; carefully re-tweaking the sparse re-training can be its important booster. 



\subsection{Ablation Study and Visualization}

\begin{wrapfigure}{r}{5cm}
\vspace{-1.2cm}
\includegraphics[width=5cm]{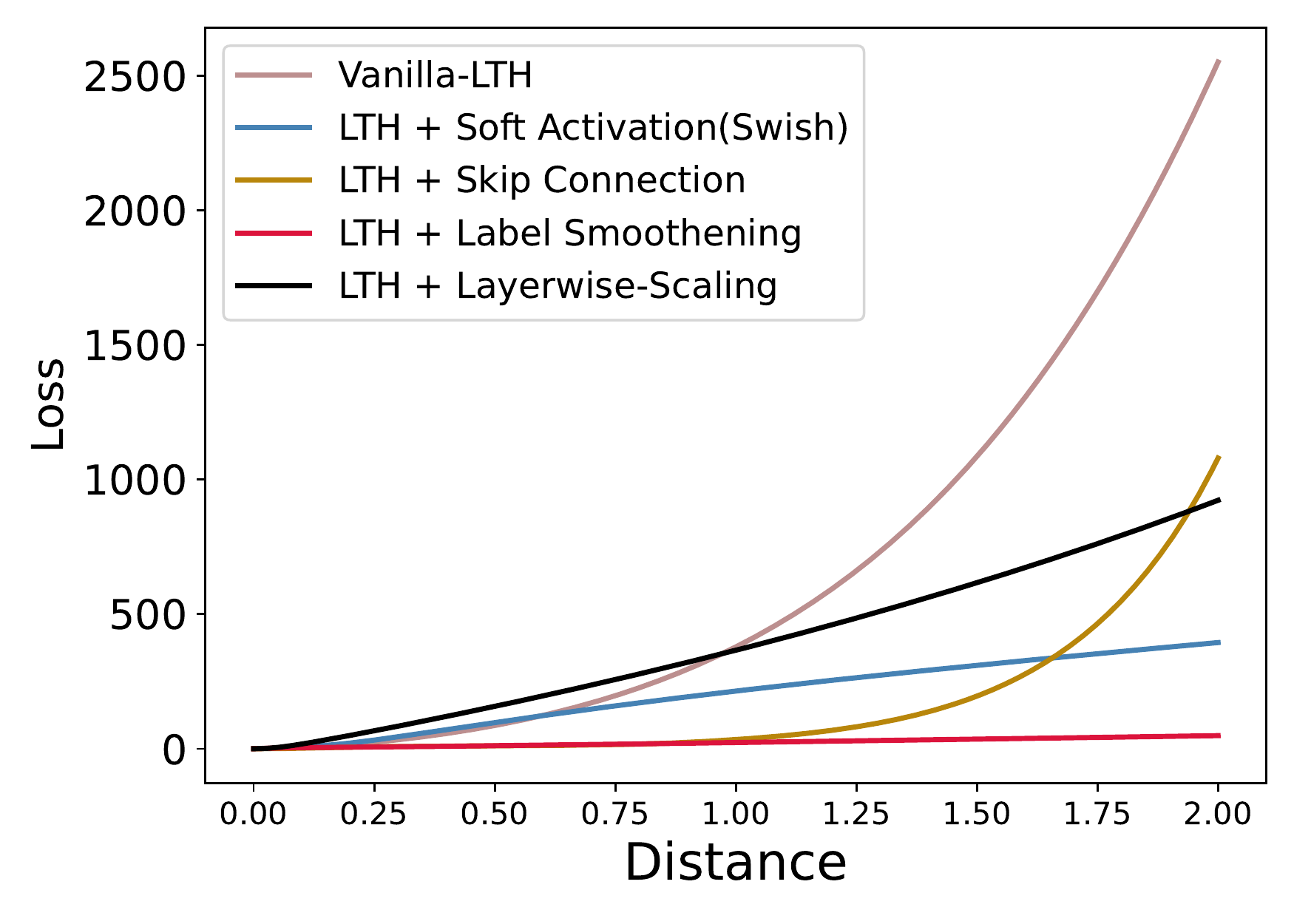}
\vspace{-0.8cm}
\caption{The change in testing loss as a function of perturbed weight distance, in the direction of top eigenvector of Hessian matrix.}
\vspace{-0.5cm}
\label{fig:commodity}
\end{wrapfigure}

\paragraph{Naive LS verses learned logit smoothening} As KD can be viewed as a learned version of label smoothening (LS) \citep{kd_and_ls}, we tried to quantify whether there is any benefit of using KD \citep{Hinton2015DistillingTK} compared to naive LS \citep{szegedy2016rethinking} in our proposed LTH (+AT and +TRT). Figure \ref{fig:LSvsKD} illustrates the benefits of replacing naive LS with KD (dense network serve as teacher). We found that KD has marginal benefits when it replaces naive LS in TRT of our proposed LTH. Also, the usage of KD leads to strong entwine between dense and sparse training, as well as an increase in computational overhead. Hence, we have conducted our experiments using naive LS by default.

\begin{figure}
\centering
\includegraphics[width=14cm]{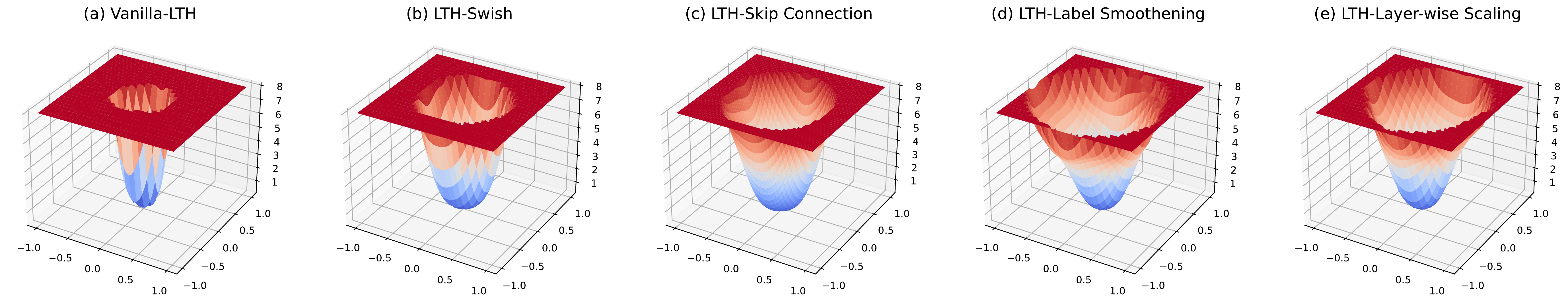}
\caption{Comparison of loss landscape of models with 91\% sparsity trained using vanilla-LTH and our Smoothness-aware techniques \textit{(architecture tweaking and training recipe tweaking)}. Loss plots are generated with the same original images randomly chosen from CIFAR-100 test dataset using \citep{loss-landscape}. z-axis denote the loss value which has been clamped at 8.0 for better visualization. }
\label{fig:loss_landscape}
\end{figure}

\paragraph{Comparison of loss landscapes}
We expect our proposed techniques to find flatter minima for sparse subnetworks training to improve its generalization, and we show it happen by visualizing the loss landscape w.r.t. both input and the weight space. Figure \ref{fig:loss_landscape} shows that our methods notably flatten the landscape w.r.t. input space, compared to vanilla-LTH baseline. Figure  \ref{fig:commodity}  follows \citep{Yao2020PyHessianNN} to perturb the trained model (91\% sparsity) in weight space in direction of the top eigenvector and show how testing loss changes with perturbation distance. Our methods present better around the achieved local minima, which suggests improved generalization.

\section{Conclusion}

This paper makes a contrary argument to the common wisdom in LTH that sparse re-training has to stick to the protocol of dense network training. We demonstrated that purposely re-tweaking the network architecture, initialization, and training protocol from dense training can actually improve spare subnetwork performance. We present two groups of techniques - architecture tweaking and training recipe tweaking, both motivated by injecting smootheness in training sparse subnetworks. Our extensive experiments across several datasets and architectures present the generalization capability of our techniques to achieve SOTA performance. Our future work will aim for more theoretical understanding of the role of our techniques in sparse (re-)training performance improvement.   

\clearpage

\bibliography{Smooth_LTH}
\bibliographystyle{iclr2022_conference}

\clearpage

\appendix
\renewcommand{\thepage}{A\arabic{page}}  
\renewcommand{\thesection}{A\arabic{section}}   
\renewcommand{\thetable}{A\arabic{table}}   
\renewcommand{\thefigure}{A\arabic{figure}}

\section{More Experiment Results} \label{sec:more_results}

\subsection{Analysis of activation sparsity of Relu and Smooth neurons}
Sparse neural networks suffer from very high activation sparsity due to the non-smooth nature of ReLU. ReLU gradient changes suddenly around zero, and this causes high proportions of network activations to be clamped to zero. Table \ref{tab:activation_sparsity} provide the activation sparsity of ReLU at different sparsity level for ResNet-18 model. Activation sparsity of the network increases with the depth for ReLU activation. When ReLU is replaced with smooth activation (Swish), the activation sparsity of the network drops below $1\%$. Soft-activations allows sparse NN training to get rid of zero effect of ReLU (zero activations) and facilitate smoother and healthy gradient flow in the network which may help the performance of sparse NN training.  
\begin{table}[h]
\caption{Activation sparsity of different layers of ResNet-18 trained on CIFAR-100 using ReLU and Swish at various sparsity levels.}
\label{tab:activation_sparsity}
\begin{center}
\begin{tabular}{l|llllllll}
\toprule
\multirow{2}{*}{Sparsity Level} & \multicolumn{2}{c}{Layer 1} & \multicolumn{2}{c}{Layer 2} & \multicolumn{2}{c}{Layer 3} & \multicolumn{2}{c}{Layer 4}\\

& ReLU & Swish& ReLU & Swish& ReLU & Swish& ReLU & Swish\\
\midrule
20\%& 28.93\%	&0.33\%	&49.23\%	&0.25\%	&62.36\%	&0.22\%	&65.50\%	&0.16\%\\
36\%&	32.29\%&	0.31\%&	47.46\%&	0.25\%&	60.40\%&	0.23\%&	65.69\%&	0.16\%\\
\midrule
67\%&	35.18\%&	0.29\%&	46.29\%&	0.25\%&	58.01\%&	0.23\%&	64.19\%&	0.16\%\\
74\%&	35.70\%&	0.31\%&	50.50\%&	0.25\%&	56.30\%&	0.23\%&	64.47\%&	0.16\%\\
\midrule
96\%&	27.43\%&	0.30\%&	38.26\%&	0.26\%&	39.78\%&	0.27\%&	56.06\%&	0.23\%\\
97\%&	27.99\%&	0.39\%&	36.63\%&	0.24\%&	39.59\%&	0.28\%&	53.18\%&	0.24\%\\
\bottomrule
\end{tabular}
\end{center}
\end{table}

\subsection{Hessian Visualization during Sparse NN Retraining}
\begin{figure}[h]
\centering
\includegraphics[width=12cm]{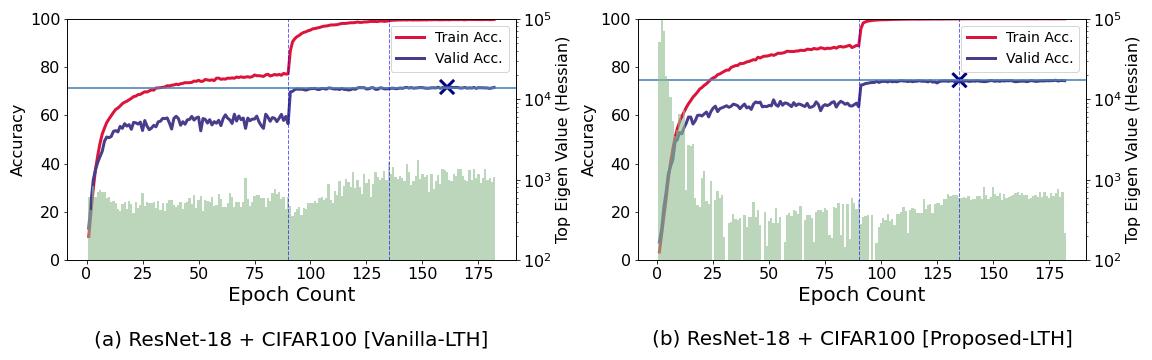}
\caption{Comparison of top eigenvalue of Hessian (mean across 3 batches every epoch) during training of our proposed LTH (+AT and +TRT) and vanilla-LTH (91\% sparsity). Left y-axis denote accuracy and Right y-axis denote top eigen value of Hessian in log scale. Dotted vertical line denote the epoch when learning rate is scaled by 0.1 and solid horizontal line denote test accuracy. Cross-sign indicate the epoch when best validation accuracy is achieved. The eigenvalues of hessian is calculated using \citep{Yao2020PyHessianNN}.}
\label{fig:hessian_train}
\end{figure}
Top eigenvalues of hessian can be used to analyze the toplogy of the loss landscape (i.e., curvature information) and understand the behaviour of different models/optimizers. Figure \ref{fig:hessian_train} presents the top eigenvalues of hessian our proposed LTH and vanilla-LTH (91\% sparsity) for each training epoch. We can observe that in initial phase of sparse subnetwork training, our methods provide high push (high eigenvalue) and soon land up in a smoother regime (low eigenvalue). Due to the initial push, our method possibly achieve higher train/val accuracy in initial phase of training (before learning rate scaling) in compare to vanilla-LTH, and ultimately a significant improvement of $+3.11\%$ in test accuracy. Note that during final stage of training, our proposed LTH have high low eigenvalue than vanilla-LTH, which indicate the presence of high smoothness.

\subsection{Generalization across OMP}

\begin{wrapfigure}{r}{6.5cm}
\vspace{-1.4cm}
\includegraphics[width=6.5cm]{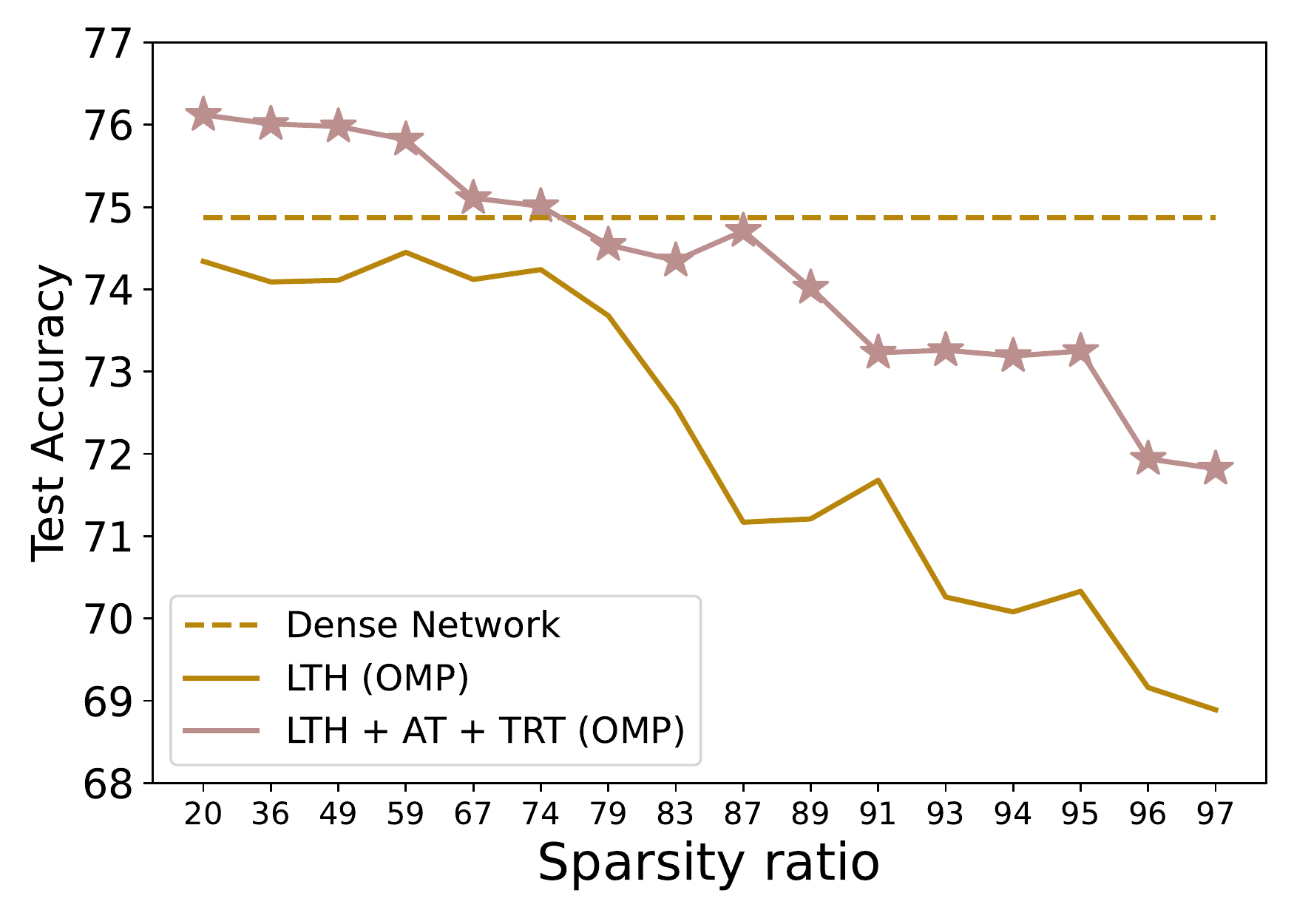}
\vspace{-0.9cm}
\caption{Testing accuracy of ResNet-18 with our proposed LTH (+AT and +TRT) when IMP is replaced with OMP.}
\vspace{-0.5cm}
\label{fig:omp}
\end{wrapfigure} 
We investigated the effectiveness of our proposed techniques when \textit{iterative magnitude pruning} is replaced with \textit{one-shot magnitute pruning} (OMP). Figure \ref{fig:omp} illustrate the generalization of the benefits of our proposed techniques when IMP is replaced with OMP. We observe that our techniques still perform significantly better than vanilla-LTH when IMP is replaced with OMP. 
\end{document}